\pdfoutput=1
\documentclass[sigconf]{acmart}

\AtBeginDocument{%
  }

\setcopyright{acmcopyright}
\copyrightyear{2023}
\acmYear{2023}
\setcopyright{rightsretained}
\acmConference[MM '23] {Proceedings of the 31st ACM International Conference on Multimedia}{October 29--November 3, 2023}{Ottawa, ON, Canada.}
\acmBooktitle{Proceedings of the 31st ACM International Conference on Multimedia (MM '23), October 29--November 3, 2023, Ottawa, ON, Canada}
\acmPrice{15.00}
\acmISBN{979-8-4007-0108-5/23/10}
\acmDOI{10.1145/3581783.3611909}
\usepackage{graphicx}
\usepackage{float}
\usepackage{subfig}
\usepackage{multirow}
\usepackage{makecell}

\begin{document}

\title{FourLLIE: Boosting Low-Light Image Enhancement by Fourier Frequency Information}

\author{Chenxi Wang}
\affiliation{%
  \institution{Sun Yat-sen University}
  \city{Shenzhen}
  \state{Guangdong}
  \country{China}
}
\email{wangchx67@mail2.sysu.edu.cn}
\author{Hongjun Wu}
\affiliation{%
  \institution{Sun Yat-sen University}
  \city{Shenzhen}
  \state{Guangdong}
  \country{China}
}
\email{wuhj33@mail2.sysu.edu.cn}

\author{Zhi Jin}
\affiliation{%
  \institution{Sun Yat-sen University}
  \city{Shenzhen}
  \state{Guangdong}
  \country{China}
}
\email{jinzh26@mail2.sysu.edu.cn}
\authornote{Corresponding author.}


\begin{abstract}
Recently, Fourier frequency information has attracted much attention in Low-Light Image Enhancement (LLIE). Some researchers noticed that, in the Fourier space, the lightness degradation mainly exists in the amplitude component and the rest exists in the phase component. By incorporating both the Fourier frequency and the spatial information, these researchers proposed remarkable solutions for LLIE. In this work, we further explore the positive correlation between the magnitude of amplitude and the magnitude of lightness, which can be effectively leveraged to improve the lightness of low-light images in the Fourier space. Moreover, we find that the Fourier transform can extract the global information of the image, and does not introduce massive neural network parameters like Multi-Layer Perceptrons (MLPs) or Transformer. To this end, a two-stage Fourier-based LLIE network (FourLLIE) is proposed. In the first stage, we improve the lightness of low-light images by estimating the amplitude transform map in the Fourier space. In the second stage, we introduce the Signal-to-Noise-Ratio (SNR) map to provide the prior for integrating the global Fourier frequency and the local spatial information, which recovers image details in the spatial space. With this ingenious design, FourLLIE outperforms the existing state-of-the-art (SOTA) LLIE methods on four representative datasets while maintaining good model efficiency. Notably, compared with a recent Transformer-based SOTA method SNR-Aware, FourLLIE reaches superior performance with only 0.31$\%$ parameters. Code is available at \hyperref[]{https://github.com/wangchx67/FourLLIE}. 
\end{abstract}



\maketitle

\section{Introduction}

The fast development of imaging devices eases image capture during our daily life. However, images obtained under low-light conditions easily suffer from unpleasing visibility, which may be unfriendly to users and some high-level computer vision tasks (e.g., action recognition \cite{pan2021view}, face detection \cite{yang2019joint}, and object detection \cite{ren2015faster}). Thus, Low-Light Image Enhancement (LLIE), which aims to recover hidden information and improve the quality of low-light images, is posed as an active research field in computer vision. 

There are many impressive LLIE methods have been proposed with great performance. Generally, they can be roughly divided into non-learning based methods \cite{jobson1997multiscale,rahman2004retinex,pizer1987adaptive,guo2016lime,fu2016fusion,wang2013naturalness,fu2016weighted} and learning based methods \cite{li2021low,lore2017llnet,wei2018deep,zhang2019kindling,jiang2021enlightengan,guo2020zero,ma2022toward,liu2021retinex,wu2022lightweight}. However, as mentioned in \cite{huang2022deep}, most of them are based on the spatial information and rarely consider the Fourier frequency information, which has been proved to be effective for improving the image quality.

Recently, some methods \cite{huang2022deep,li2023embedding} explore the Fourier frequency information for LLIE. They find that, in the Fourier space, the most lightness representation is concentrated in the amplitude component while the phase component contains the lightness-irrelevant information (e.g., structure \cite{huang2022deep} or noise \cite{li2023embedding}). Based on this observation, they integrate both the Fourier frequency and spatial information into neural networks and achieve impressive results.

\begin{figure*}
    \centering
    \includegraphics[width=1.\textwidth]{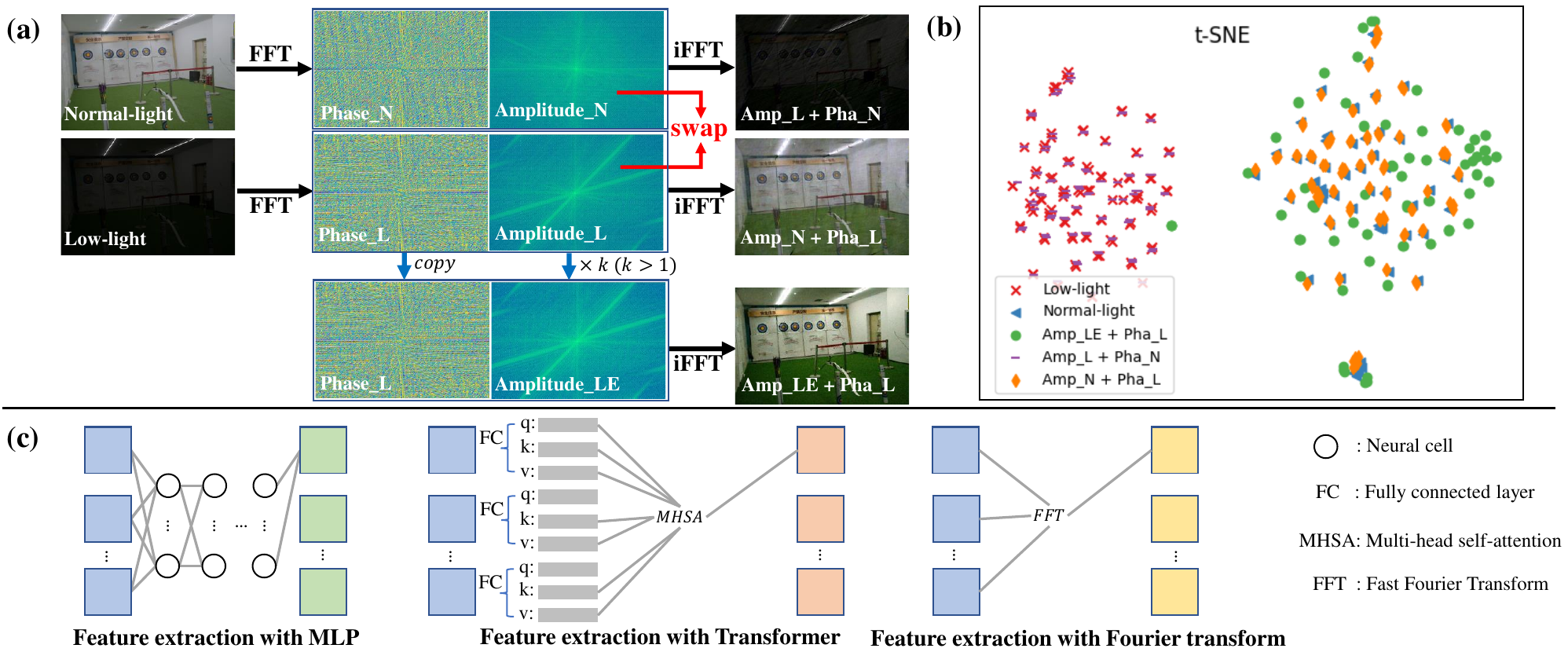}
     \vspace{-0.3cm}
    \caption{Our motivations. (a) The amplitude components of low-light and normal-light images with the same context are swapped to get recombined results Amp$\_$L+Pha$\_$N and Amp$\_$N+Pha$\_$L (the top and middle rows). The amplitude component of the low-light image is enlarged by a constant value $k$ to get Amp$\_$LE+Pha$\_$L (the bottom row). From visual results we can infer that the amplitude component reflects the lightness representation of an image and the magnitudes of amplitude represent the magnitudes of lightness. (b) The t-SNE \cite{van2008visualizing} embeddings of low-light, normal-light, Amp$\_$LE+Pha$\_$L, Amp$\_$L+Pha$\_$N and Amp$\_$N+Pha$\_$L. The distributions of   Amp$\_$LE+Pha$\_$L and Amp$\_$N+Pha$\_$L are similar to normal-light, while the distribution of Amp$\_$L+Pha$\_$N is similar to low-light. (c) The features extracted by Transformer \cite{vaswani2017attention}, MLP, and Fourier transform  \cite{brigham1967fast} are global information. However, the feature extracted by the Fourier transform does not introduce massive parameters of neural networks.}
    \label{fig:motivations}
    \vspace{-0.4cm}
\end{figure*}

Inspired by previous Fourier-based works \cite{huang2022deep,li2023embedding}, we further explore the properties of the Fourier frequency information for LLIE. Our motivations are shown in Fig. \ref{fig:motivations}. Firstly, given two images with the same context but different light conditions (i.e., low-light and normal-light), we swap their amplitude components and combine them with corresponding phase components in the Fourier space. The recombined results show that the light conditions are swapped following the amplitude swapping (see the top two rows of Fig. \ref{fig:motivations} (a)). This phenomenon indicates \textbf{the amplitude component represents the lightness of an image}. Then, by only enlarging the magnitude of the amplitude component of the low-light image and keeping the phase component, we find that the low-light image is brightened (see the button row of Fig. \ref{fig:motivations} (a)) and has closer distributions to the normal-light image (see Fig. \ref{fig:motivations} (b)). From this, we conclude that \textbf{the lightness of low-light images can be improved by enlarging the magnitude of its amplitude component in the Fourier space}. Moreover, we notice that both Transformer \cite{vaswani2017attention}, MLP, and \textbf{Fourier transform \cite{brigham1967fast} can extract global information} (see Fig. \ref{fig:motivations} (c)), while feature extraction with the Fourier transform is more efficient since \textbf{it does not introduce massive parameters of neural networks}.

Based on the above observations, in this work, we propose a new LLIE method (called FourLLIE) by leveraging the Fourier frequency information. FourLLIE accomplishes enhancement through two stages: frequency stage and spatial stage. In the frequency stage, FourLLIE achieves lightness improvement by estimating the transform map of the amplitude component. This process is similar to illumination map estimation methods \cite{liu2021retinex,ma2022toward,wang2019underexposed} based on Retinex theory \cite{jobson1997multiscale}, while it is implemented in the Fourier space. Then, we introduce the Signal-to-Noise-Ratio (SNR) map into the spatial stage for further details refinement. Generally, the image regions with different SNR values indicate different amount of information and noise. The regions with higher SNR values are better improved with local information, oppositely, they are better improved with global information \cite{xu2022snr}. Based on this property and the global property of the Fourier frequency information, we propose to recover image regions with higher SNR values by the local spatial information and image regions with lower SNR values by the Fourier frequency information. In this way, a low-light image is first brightened in the frequency stage and then refined in the spatial stage.

Compared with the existing Fourier-based image enhancement methods \cite{huang2022deep,li2023embedding}, we go step further to explore the property that the magnitude of amplitude is positively correlated with the magnitude of lightness, and estimate a transform map to improve the magnitude of the amplitude component. Experiments in Sec. \ref{frequency_stage} demonstrate the effectiveness of the proposed design. Besides, we introduce the SNR map to well utilize the global property of the Fourier frequency information. Compared with SNR-Aware \cite{xu2022snr}, which adopts Transformer \cite{vaswani2017attention} as the global information extractor to process low SNR regions, we adopt the Fourier frequency information to represent the global information and remarkably reduce model complexities (see experiments in Sec. \ref{spatial_stage}). To sum up, our contributions are three folds: 

\textbf{1)} We explore the positive correlation between the magnitudes of amplitude component in the Fourier space and lightness in the spatial space. By utilizing this property, we provide an effective and reasonable way to achieve lightness improvement in the Fourier space. 

\textbf{2)} We introduce the SNR map to integrate the Fourier frequency information and spatial information. It takes full advantage of the global property of the Fourier frequency information.

\textbf{3)} We conduct extensive experiments on four commonly used LLIE datasets, demonstrating proposed method outperforms existing SOTA methods while preserving good model efficiency.

\begin{figure*}
    \centering
    \includegraphics[width=1.\textwidth]{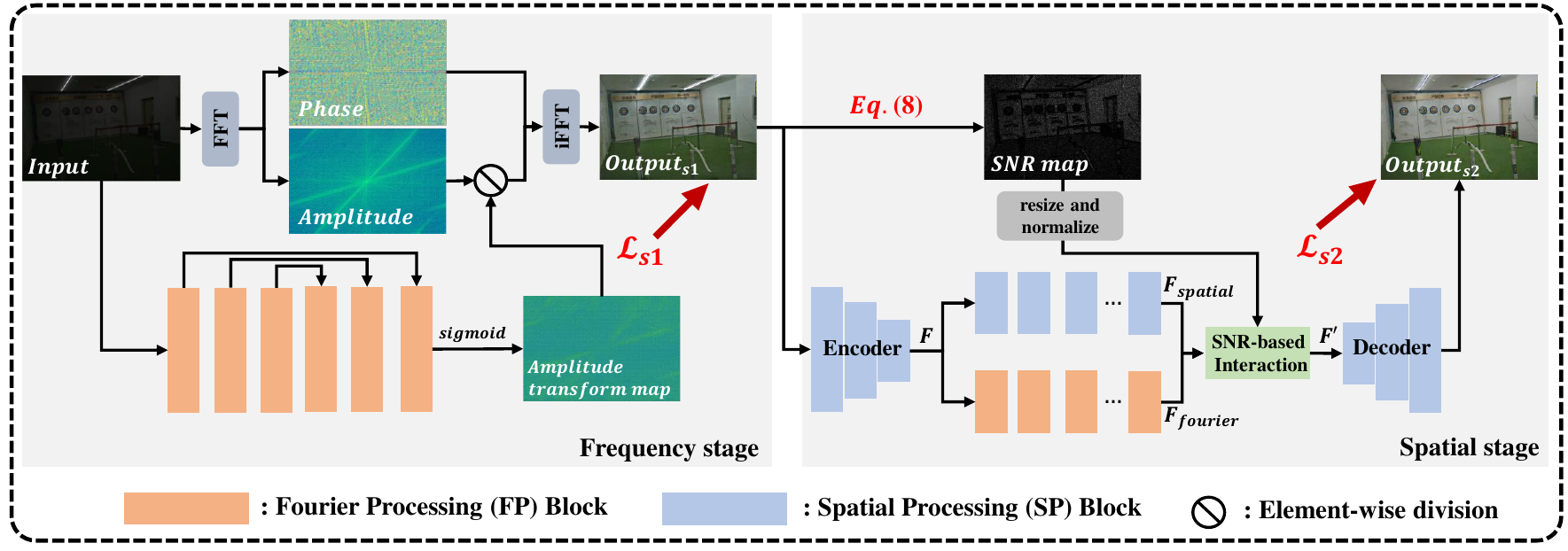}
     \vspace{-0.3cm}
    \caption{Overall architecture of the proposed method. The input image is first processed in the frequency stage to improve the lightness. Then, an SNR-based spatial stage is designed to further recover the details. Note that, in the spatial stage, the SP blocks in encoder and decoder modules are followed by down-sampling and up-sampling operations.}
    \label{fig:pipeline}
\end{figure*}

\section{Related Work}

\subsection{Low-Light Image Enhancement}
Low-Light Image Enhancement (LLIE) aims to improve the quality of low-light images. Before the development of deep learning, many non-learning based LLIE methods, which include HE-based \cite{pizer1987adaptive} and Retinex-based \cite{jobson1997multiscale,rahman2004retinex}, are proposed. However, these methods may hardly handle noise and color well. Recently, with the rapid development of deep learning, a rich set of learning based architectures \cite{li2021low,lore2017llnet,wei2018deep,zhang2019kindling,jiang2021enlightengan,guo2020zero,ma2022toward,liu2021retinex,xu2022snr,zheng2022enhancement,jin2020dual,jin2019flexible,pan2022chebylighter,liang2022learning} are proposed. Lore \emph{et al.} \cite{lore2017llnet} first introduced Convolutional Neural Networks (CNNs) to LLIE. Gharbi \emph{et al.} \cite{gharbi2017deep} accomplished enhancement by the bilateral up-sampling network. Zamir \emph{et al.} \cite{zamir2020learning} proposed more advanced CNN architecture, which includes attention mechanisms and multi-scale designs, to achieve robust LLIE. Xu \emph{et al.} \cite{xu2022snr} introduced Transformer \cite{vaswani2017attention} and SNR map to LLIE. Besides, another line of LLIE methods tries to combine Retinex theory and deep learning. Wei \emph{et al.} \cite{wei2018deep} and Zhang \emph{et al.} \cite{zhang2019kindling} achieved enhancement by decomposing a low-light image to reflectance and illumination components. Methods in \cite{wang2019underexposed, liu2021retinex, ma2022toward} estimated the illumination map of low-light images. However, the above methods may pay less attention to the Fourier frequency information, which is effective for LLIE.

Recently, researchers explore the Fourier frequency information for LLIE. Huang \emph{et al.} \cite{huang2022deep} pointed out amplitude component can reflect lightness representation of under-/over- exposure images. Li \emph{et al.} \cite{li2023embedding} found lightness and noise can be decomposed in the Fourier space. Based on these observations, they reached remarkable performance by integrating both the Fourier frequency information and spatial information. However, these works may hardly explore the relationship between the amplitude components of low-/normal- light images and well utilize the global property of the Fourier frequency information.

\subsection{Fourier Frequency Information}
Fourier frequency information has been demonstrated as the effective representation in many computer vision areas \cite{Yang_2020_CVPR,xu2021fourier,jiang2021focal,zhou2022deep,fuoli2021fourier,yu2022frequency,huang2022deep,li2023embedding,zhou2022spatial,zhou2022adaptively,guo2022exploring}. For example, Xu \emph{et al.} \cite{xu2021fourier} developed a Fourier-based data augmentation for domain generation. Fuoli \emph{et al.} \cite{fuoli2021fourier} adopted Fourier-based loss to help restore the high-frequency information in image super-resolution. Yu \emph{et al.} \cite{yu2022frequency} applied the Fourier frequency information to the image dehazing and methods in \cite{zhou2022spatial,zhou2022adaptively} utilized it for pan-sharpening. Zhou \emph{et al.} \cite{zhou2022deep} designed a Fourier-based up-sampling and can improve many computer vision tasks in a plug-and-play way. Besides, Huang \emph{et al.} \cite{huang2022deep} and Li \emph{et al.} \cite{li2023embedding} developed the Fourier-based LLIE algorithms, while they still have certain limitations as analyzed before. The above advances provide wide applications of the Fourier frequency information and motivate us to further explore its properties in LLIE.

\section{Method}

\subsection{Fourier Frequency Information}
Firstly, we briefly introduce the Fourier frequency information. Given an input image $x$, whose shape is $H \times W$, the transform function $\mathcal{F}$ which converts $x$ to the Fourier space $X$ can be represented as follow:
\begin{equation}\label{FFT}
    \mathcal{F} (x)(u,v) = X(u,v) = \frac{1}{\sqrt{HW} }\sum_{h=0}^{H-1} \sum_{w=0}^{W-1}  x(h,w)e^{-j2\pi(\frac{h}{H}u+\frac{w}{W}v} )  
\end{equation}
where $h$, $w$ are the coordinates in the spatial space and $u$, $v$ are the coordinates in the Fourier space, $j$ is the imaginary unit, the inverse process of $\mathcal{F}$ is denoted as $\mathcal{F}^{-1}$, $X(u,v)$ consists of complex values and can be represented by:
\begin{equation}\label{FFT}
    X(u,v) = R(X(u,v))+jI(X(u,v))
\end{equation}
where $R(X(u,v))$ and $I(X(u,v))$ are the real and imaginary parts of $X(u,v)$, respectively.

Besides, $X(u,v)$ in the Fourier space can be represented by an amplitude component $\mathcal{A}(X(u,v))$ and a phase component $\mathcal{P}(X(u,v))$ as follows:
\begin{equation}
    \mathcal{A}(X(u,v)) = \sqrt[]{R^2(X(u,v))+I^2(X(u,v))} 
\end{equation}
\begin{equation}
    \mathcal{P}(X(u,v)) = arctan[\frac{I(X(u,v)}{R(X(u,v)} ]
\end{equation}
where the $R(X(u,v))$ and $I(X(u,v))$ can also be obtained by:
\begin{equation}
    \begin{split}
    R(X(u,v)) = \mathcal{A}(X(u,v)) \times cos(\mathcal{P}(X(u,v))) \\
    I(X(u,v)) = \mathcal{A}(X(u,v)) \times sin(\mathcal{P}(X(u,v))) 
    \end{split}
\end{equation}

\begin{figure}
    \centering
    \includegraphics[width=1.\columnwidth]{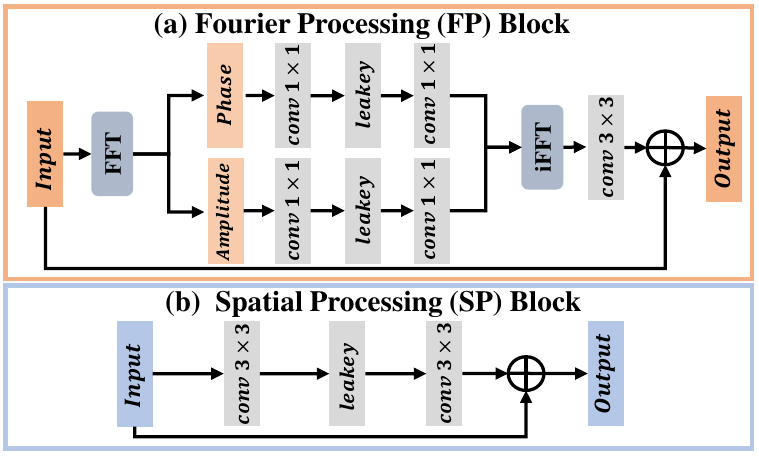}
     \vspace{-0.3cm}
    \caption{The illustration of Fourier Processing (FP) block and Spatial Processing (SP) block.}
    \label{fig:processBlock}
    \vspace{-0.4cm}
\end{figure}

According to previous Fourier-based methods \cite{huang2022deep,li2023embedding} and Fig. \ref{fig:motivations}, we conclude that: \textit{1) the lightness of the low-light image can be improved by enlarging the magnitude of amplitude component in the Fourier space. 2) the Fourier transform can extract global information and does not introduce massive parameters of neural networks.} Based on the above two conclusions, we propose to enhance low-light images in two stages: frequency stage and spatial stage. The frequency stage improves the lightness of the low-light image by estimating the amplitude transform map to improve the magnitude of the amplitude component in the Fourier space. The spatial stage utilizes the global properties of the Fourier frequency information and SNR map to further recover the details. The detailed implementations are in the following parts.

\subsection{Frequency Stage in FourLLIE \label{frequency_stage}}

The enhancement in the Frequency stage is shown in Fig. \ref{fig:pipeline}. The input image is fed into six Fourier Processing (FP) blocks with skip connections to estimate the amplitude transform map. FP block is designed to extract the Fourier frequency features (as shown in Fig. \ref{fig:processBlock} (a)). The input of each FP block is first transformed to the Fourier space to get the amplitude and phase components. Then, for each component, two $1\times1$ convolutional layers with a LeakyReLU activation are applied to extract features. Finally, these two components are transformed back to the spatial space followed by a $3\times3$ convolutional layer and the residual of input. The final $sigmoid$ activation limits the transform map within $(0,1)$. Therefore, the result of the frequency stage $Output_{s1}$ is obtained by the division between the amplitude component of the input image and estimated transform map. This process can be expressed as:
\begin{equation}
\begin{split}
A_{in} = \mathcal{A}&(\mathcal{F}(input)), \quad P_{in} = \mathcal{P}(\mathcal{F}(input))\\
&A_{out} = A_{in} / (M + \epsilon ) \\
R_{out} = A_{out} &\times cos(P_{in}) \quad I_{out} = A_{out} \times sin(P_{in})\\
&Output_{s1} = \mathcal{F}^{-1}(R_{out} + jI_{out})
\end{split} 
\end{equation}
where $A_{in}$ and $P_{in}$ are amplitude and phase components of the input image, respectively, $A_{out}$ is the output amplitude component, $R_{out}$ and $I_{out}$ represent the real and imaginary parts of $Output_{s1}$ in the Fourier space, respectively, $M$ is estimated amplitude transform map and $\epsilon$ is set to $1.0 \times 10^{-8}$ for avoiding zero-division.

\begin{figure}
    \centering
    \includegraphics[width=1.\columnwidth]{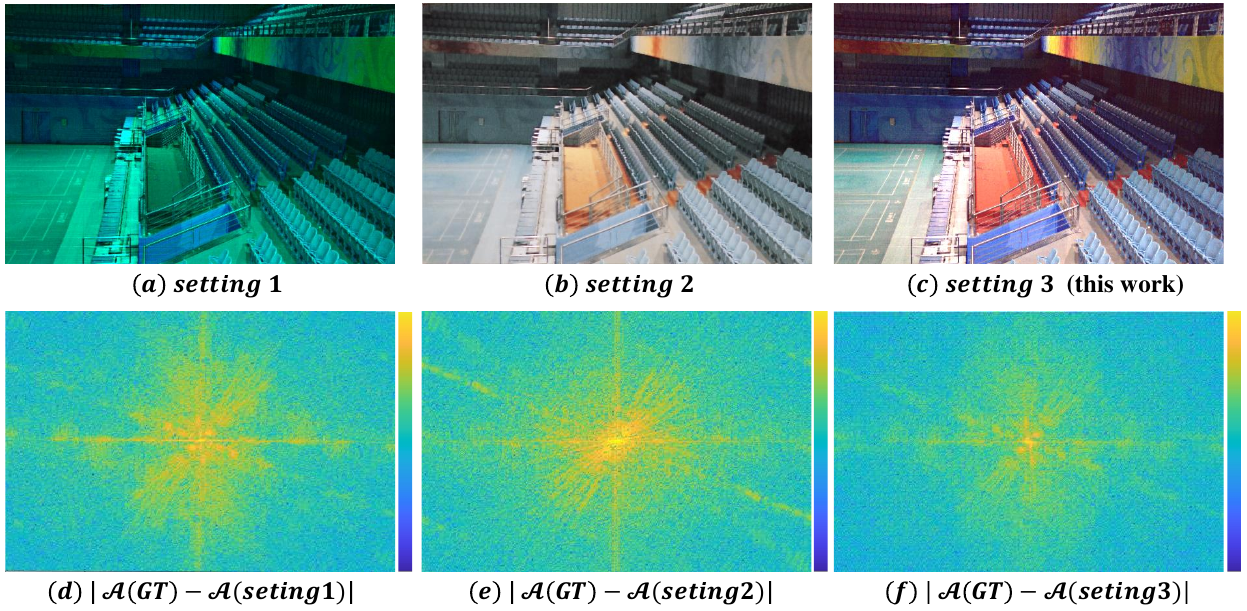}
     \vspace{-0.3cm}
    \caption{Ablation study for amplitude prediction ways. GT represents the ground truth image and $\mathcal{A}(.)$ represents to extract amplitude component in the Fourier space. (a), (b), and (c) present the visual results of different settings. (d), (e), and (f) present the error maps of amplitude components between different settings and GT. It can be seen that the setting of this work (setting 3) reaches the best results no matter in the spatial space or Fourier space. }
    \label{fig:importanceOFmap}
    \vspace{-0.4cm}
\end{figure}

In this way, the lightness of the input low-light image is improved with the enlargement of its amplitude. The loss involved in the frequency stage $\mathcal{L}_{s1}$ is expressed as:
\begin{equation}
\mathcal{L}_{s1} = \left | \left | \mathcal{A}(\mathcal{F}(Output_{s1}))-\mathcal{A}(\mathcal{F}(GT)) \right |  \right | _{2}
\end{equation}
where GT is the ground truth image.

Compared with FECNet \cite{huang2022deep}, which also improves the lightness by applying a constrain on the amplitude component, we utilize the positive correlation (an amplitude transform map within (0,1)) between the magnitudes of the amplitude and lightness. It makes enhancement in the Fourier space more reasonable and robust for LLIE. 

To verify the effectiveness of estimating the amplitude transform map, we conduct an experiment with three different settings: \textbf{1)} estimating the output amplitude directly and applying a constraint on the estimated amplitude component, \textbf{2)} estimating the
output image in the spatial space and applying a constraint on the amplitude component by transforming the obtained image to the Fourier space (setting of FECNet \cite{huang2022deep}), \textbf{3)} estimating the amplitude transform map and applying a constraint on the amplitude component transformed by estimated transform map (setting of this work). The detailed experiment settings can be found in the supplementary material. We denote them as setting 1, setting 2, and setting 3 for simplicity. 

As shown in Fig. \ref{fig:importanceOFmap} (a) and (d), since the amplitude component is very complex and does not have structural property or other related priors like images in the spatial space, it is hard to learn a well enhanced amplitude directly by a neural network. As for setting 2, it easily leads to detail losses (e.g., color) for low-light images since it neglects the relationship between amplitude components of low-light and normal-light images (see Fig. \ref{fig:importanceOFmap} (b) and (e)). However, when predicting the amplitude component by estimating the transform map like setting 3, the details of the image and structure of amplitude can be recovered better as shown in  Fig. \ref{fig:importanceOFmap} (c) and (f).
\begin{figure}
    \centering
    \includegraphics[width=1.\columnwidth]{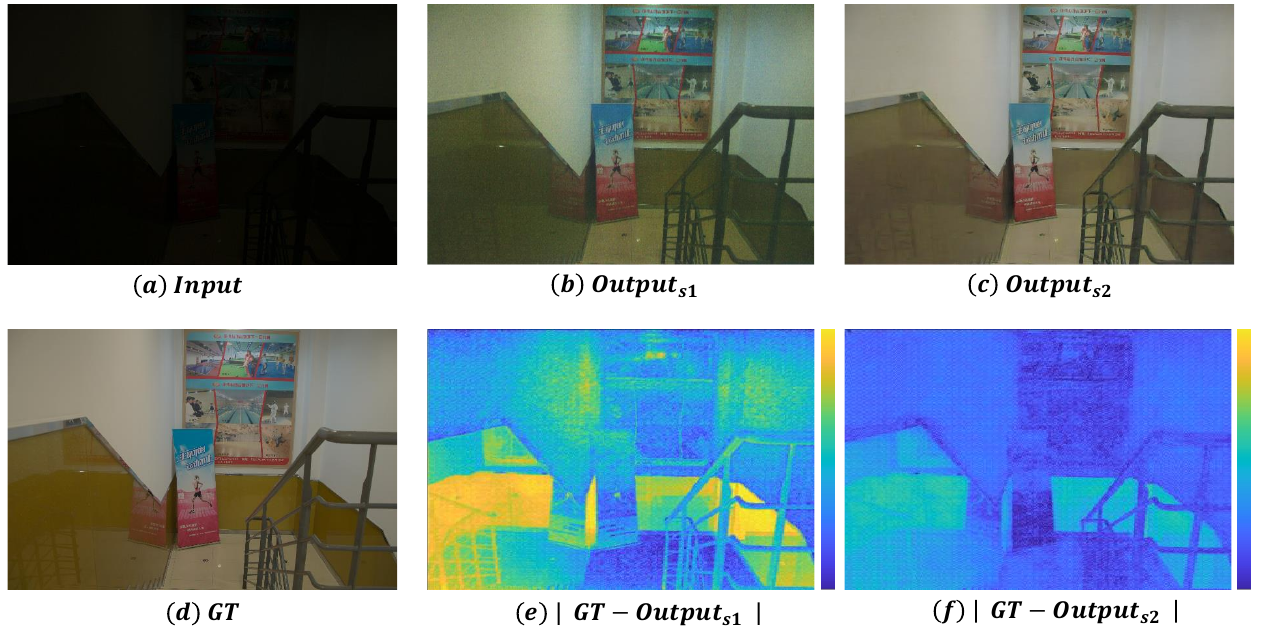}
     \vspace{-0.4cm}
    \caption{Visualization of outputs in different stages. (b) and (c) are the outputs of the frequency stage and the spatial stage, respectively. (e) and (f) present the error maps between the outputs of different stages and GT. It can be seen that the frequency stage improves the lightness and the spatial stage recover the details. }
    \label{fig:twoStage_visual}
    \vspace{-0.5cm}
\end{figure}

\subsection{Spatial Stage in FourLLIE \label{spatial_stage}}

The lightness of the low-light image can be improved well in the frequency stage, however, it still suffers some detail degradation (as shown in Fig. \ref{fig:twoStage_visual} (b) and (e)). Meanwhile, based on the property of SNR map, regions of an image with lower SNR values need long-range (global) operations to restore, while the regions with higher SNR values prefer short-range (local) operations. In this work, we further extend that the regions with lower SNR values tend to be processed in the Fourier space, while the regions with higher SNR values prefer to be processed in the spatial space. Therefore, we propose to integrate the global Fourier frequency information and local spatial information based on the SNR map and recover details. The enhancement in the spatial stage is designed as shown in Fig. \ref{fig:pipeline}. Given the output of the frequency stage $Output_{s1}$, we first compute the SNR map $S$ followed \cite{xu2022snr} as:

\begin{equation}
    \hat{I_{g}} = blur(I_{g}), \qquad N = abs(I_{g}-\hat{I_{g}}),\qquad S=\hat{I_{g}}/N
\end{equation}
where $I_{g}$ is the gray-scale version of $Output_{s1}$, $blur$ is a Gaussian blur with kernel size of  $5$, $abs(.)$ represents to compute the absolute value, $\hat{I_{g}}$ represents the noise-free version of $I_{g}$, $N$ represents the noise component. Then, we feed the $Output_{s1}$ into an encoder to extract the features $F$. The features $F$ then separately go through Fourier Processing (FP) blocks and Spatial Processing (SP) blocks to produce the global features $F_{fourier}$ and the local features $F_{spatail}$, respectively. As shown in Fig. \ref{fig:processBlock}, FP blocks process features in the Fourier space and can extract global features, while SP blocks process features in the spatial space and can extract local features. Next, the $F_{fourier}$ and $F_{spatail}$ are combined based on SNR map $S$ as:

\begin{equation}
    F'=F_{spatail}\times S+F_{fourier}\times(1-S)
\end{equation}
where $F'$ is the output features. Note that $S$ is normalized to $[0,1]$ and resized to the corresponding shape as $F_{spatail}$ and $F_{fourier}$. The result of the spatial stage $Output_{s2}$ is produced by feeding the output features $F'$ into the decoder. As shown in Fig. \ref{fig:twoStage_visual} (c) and (f), the noise contained in $Output_{s1}$ is removed after the spatial stage. The loss involved in the spatial stage $\mathcal{L}_{s2}$ can be expressed as:
\begin{equation}
\begin{split}
\mathcal{L}_{s2} = &\left | \left | Output_{s2}-GT \right |  \right | _{2} + \alpha \left | \left | \phi (Output_{s2})-\phi (GT) \right |  \right | _{2}\\
\end{split}
\end{equation}
where $\phi $ is a pre-trained VGG \cite{simonyan2014very} network, $\alpha$ is a weight factor and set to 0.1 empirically.

Compared with previous Fourier-based LLIE methods \cite{huang2022deep,li2023embedding}, which accomplish spatial-frequency interaction by features concatenation, FourLLIE introduces the SNR map as prior and well utilizes the global property of the Fourier frequency information. Compared with SNR-Aware \cite{xu2022snr}, which uses Transformer \cite{vaswani2017attention} to extract global information, Fourier-based processing can reduce model complexity significantly. 

To verify the effectiveness of using Fourier frequency information, we conduct an experiment by replacing the Transformer blocks in SNR-Aware \cite{xu2022snr} with FP blocks and training it on LSRW-Huawei \cite{hai2023r2rnet} dataset. We denote SNR-Aware with Transformer blocks as SNR-Aware-Transformer and SNR-Aware with FP blocks as SNR-Aware-Fourier. As shown in Table \ref{tab:snr_exp}, SNR-Aware-Fourier reaches competitive performance with only about 1/20 parameters of SNR-Aware-Transformer. Besides, we find SNR-Aware-Fourier is not sensitive to channels of the latent features. SNR-Aware-Fourier with only 16 channels of the latent features still can perform well on LSRW-Huawei. Nevertheless, the reduction of channels for SNR-Aware-Transformer may lead to performance degradation. We infer that it benefits from the inherently global nature of the Fourier frequency information, which does not rely on massive parameters of the neural network.

Finally, the overall loss $\mathcal{L}_{total}$ of FourLLIE can be expressed as :
\begin{equation}
    \mathcal{L}_{total} = \mathcal{L}_{s2} + \lambda\mathcal{L}_{s1}
\end{equation}
where $\lambda$ is a weight factor set to 0.01 empirically.

\begin{table}[]
\caption{Effectiveness of the Fourier frequency information. We replace Transformer blocks with FP blocks in SNR-Aware \cite{xu2022snr}. The best results are boldfaced and the second-best ones are underlined. Note that ``nc'' represents the number of channels of the latent features.}
 \vspace{-0.2cm}
    \label{tab:snr_exp}
    \resizebox{\columnwidth}{!}{
\renewcommand{\arraystretch}{1.1} 
    \centering

    \begin{tabular}{c c c c c }
     \toprule
      Methods & nc  &  PSNR & SSIM & $\#$Para  \\
     \midrule
     \multirow{3}{*}{SNR-Aware-Transformer} & 64 &\textbf{20.67}&\textbf{0.5910}& 39.12M \\
      & 32 & 20.45 & \underline{0.5892} & 12.94M\\
     & 16 & 20.02 & 0.5808 & 4.81M\\
      \midrule
     \multirow{3}{*}{SNR-Aware-Fourier} & 64 &20.62&\textbf{0.5910}&1.46M \\
      & 32 &\underline{20.63}&0.5868&\underline{0.37M}\\
     & 16 &20.61&0.5886&\textbf{0.09M}\\
     \bottomrule
    \end{tabular}
    }
\vspace{-0.3cm}
\end{table}

\begin{table*}
\caption{Quantitative comparison on the LOL-Real \cite{yang2021sparse}, LOL-Synthetic \cite{yang2021sparse}, LSRW-Huawei \cite{hai2023r2rnet}, and LSRW-Nikon \cite{hai2023r2rnet}.
The best results are boldfaced and the second-best ones are underlined.}
    \label{tab:experiments_lol}
       \resizebox{\textwidth}{!}{
\renewcommand{\arraystretch}{1.} 
    \centering
      \begin{tabular}{c c c c c c c c c c c c c c}
         \toprule

     \multirow{2}{*}{Methods} &\multicolumn{3}{c}{LOL-Real \cite{yang2021sparse}}&\multicolumn{3}{c}{LOL-Synthetic \cite{yang2021sparse}}&\multicolumn{3}{c}{LSRW-Huawei \cite{hai2023r2rnet}}&\multicolumn{3}{c}{LSRW-Nikon \cite{hai2023r2rnet}}&\multirow{2}{*}{\makecell[c]{\#Param\\(M)}} \\
      \cmidrule(r){2-4} \cmidrule(r){5-7} \cmidrule(r){8-10} \cmidrule(r){11-13}
     &    PSNR   &   SSIM  &   LPIPS  &   PSNR    &   SSIM &   LPIPS &   PSNR     &   SSIM   &   LPIPS &  PSNR   &   SSIM &   LPIPS   \\
        \midrule
      LIME \cite{guo2016lime} & 15.24 & 0.4190  &0.2203 & 16.88 & 0.7578 & 0.1041 & 17.00 & 0.3816 & 0.2069 & 13.53 & 0.3321 & 0.1272 & - \\
      MF \cite{fu2016fusion} & 18.72 & 0.5089  &0.2401 & 17.50 & 0.7737 & 0.1075 & 18.26 & 0.4283 & 0.2153 & 15.44 & 0.3997 & \underline{0.1269} & - \\
      NPE \cite{wang2013naturalness} & 17.33 & 0.4642  &0.2359 & 16.60 & 0.7781 & 0.1079 & 17.08 & 0.3905 & 0.2303 & 14.86 & 0.3738 & 0.1464 & - \\
      SRIE \cite{fu2016weighted}  & 14.45 & 0.5240  & 0.2160 & 14.50 & 0.6640 & 0.1484 & 13.42 & 0.4282 & 0.2166 & 13.26 & 0.3963 & 0.1396 & - \\
      DRD \cite{wei2018deep} & 16.08 & 0.6555  & 0.2364 & 18.28 & 0.7737 & 0.1470 & 18.23 & 0.5220 & 0.1926 & 15.18 & 0.3809 & 0.1690 & 0.86\\
      Kind \cite{zhang2019kindling} & 20.01 & 0.8412 & 0.0813& 22.62 & 0.9041 & 0.0515  & 16.58 & 0.5690 & 0.2259& 11.52 & 0.3827 & 0.1860 & 8.02\\
      Kind++ \cite{zhang2021beyond} & 20.59 & 0.8294  & 0.0875  & 21.17 &0.8814& 0.0678 & 15.43 & 0.5695 & 0.2366& 14.79 & 0.4749 & 0.2111   & 8.27\\
      MIRNet \cite{zamir2020learning} & \underline{22.11} & 0.7942 & 0.1448 & 22.52 & 0.8997 &  0.0568 & 19.98& 0.6085 & 0.2154 & 17.10 & \underline{0.5022} & 0.2170 & 31.79\\
      SGM \cite{yang2021sparse} & 20.06 & 0.8158  & \underline{0.0727} & 22.05 & 0.9089 & 0.4841 & 18.85 & 0.5991& 0.2492 & 15.73 & 0.4971 & 0.2234 & 2.31 \\
      FECNet \cite{huang2022deep} & 20.67 & 0.7952 & 0.0995 & 22.57 & 0.8938  & 0.0699 & \underline{21.09} & 0.6119 & 0.2341 & 17.06 & 0.4999 & 0.2192  & \underline{0.15}\\
      HDMNet \cite{liang2022learning} & 18.55 & 0.7132 & 0.1717 & 20.54 & 0.8539  & 0.0690 & 20.81 & 0.6071 & 0.2375 & 16.65 & 0.4870 & 0.2157  & 2.32\\
      SNR-Aware \cite{xu2022snr} & 21.48 & \textbf{0.8478} & 0.0740 & \underline{24.13} & \textbf{0.9269}  & \textbf{0.0318} & 20.67 & 0.5910 & \underline{0.1923} & \underline{17.54} & 0.4822 & \textbf{0.0982}  & 39.12\\
      Bread \cite{guo2023low} & 20.83 & 0.8217 & 0.0949 & 17.63 & 0.8376  & 0.0681 & 19.20 & \underline{0.6179} & 0.2203 & 14.70 & 0.4867 & 0.1766 & 2.12\\
     \midrule
     \multicolumn{1}{c}{Ours} & \textbf{22.34} & \underline{0.8468}  & \textbf{0.0511} & \textbf{24.65} & \underline{0.9192} & \underline{0.0389} & \textbf{21.30} & \textbf{0.6220} & \textbf{0.1719} & \textbf{17.82} & \textbf{0.5036} & 0.2150  & \textbf{0.12}\\
     \bottomrule
     
    \end{tabular}}
    \vspace{-0.4cm}
\end{table*}

\begin{figure*}
    \centering
    \includegraphics[width=1.\textwidth]{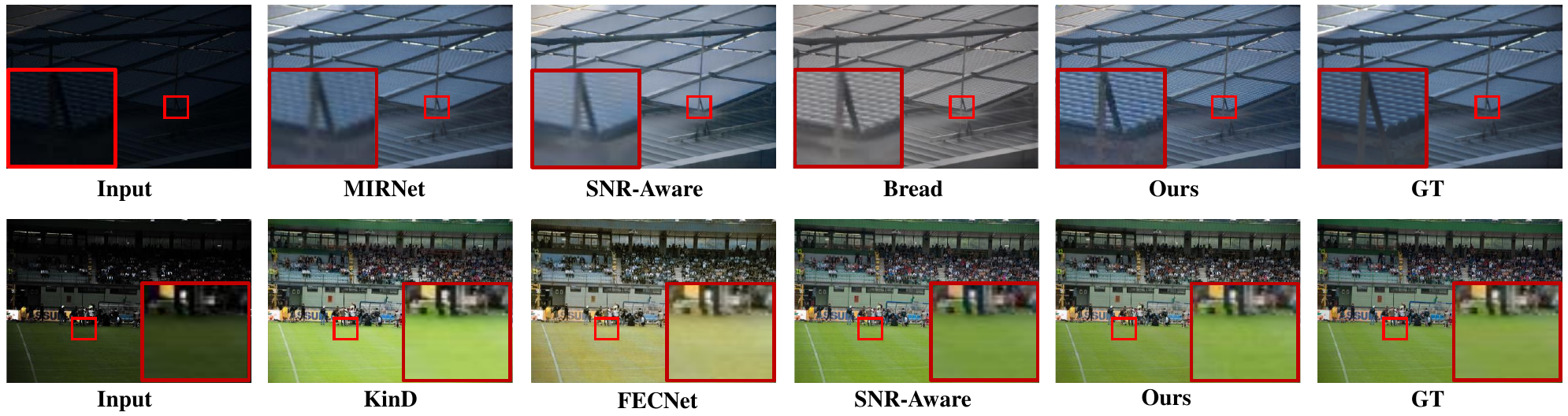}
     \vspace{-0.5cm}
    \caption{Qualitative comparison in LOL-Real \cite{yang2021sparse} (the first row) and LOL-Synthetic \cite{yang2021sparse} (the second row). It can be seen that the proposed method reaches the best visual results. Note that we compare the proposed method with three benchmarks with the best PSNR values on the corresponding datasets. }
    \label{fig:comparison_LOL}
    \vspace{-0.2cm}
\end{figure*}

\section{Experiment}

\subsection{Datasets and Implementation Details}
We choose four widely used LLIE datasets for evaluating FourLLIE, including LOL-Real \cite{yang2021sparse}, LOL-Synthetic \cite{yang2021sparse}, LSRW-Huawei \cite{hai2023r2rnet}, and LSRW-Nikon \cite{hai2023r2rnet}. LOL-Real is captured in real scenes by changing exposure time and ISO. It contains 689 low-/normal- light image pairs for training and 100 low-/normal- light image pairs for testing. It is worth noting that LOL-Real is the extended version of LOL \cite{wei2018deep}, which contains 485 training image pairs and 15 testing image pairs. Considering the image pairs from LOL-Real and LOL are overlapped, we only evaluate in LOL-Real, since it is more diverse. LOL-Synthetic is synthesized from raw images by analyzing the distribution of luminance channels of low-/normal- light images. It contains 900 low-/normal- light image pairs for training and 100 low-/normal- light image pairs for testing. LSRW-Huawei and LSRW-Nikon are captured in real scenes like LOL-Real but with different devices. LSRW-Huawei is collected by a Huawei P40 Pro and LSRW-Nikon is collected by a Nikon D7500. LSRW-Huawei contains 3150 training image pairs and 20 testing image pairs. LSRW-Nikon contains 2450 training image pairs and 30 testing image pairs. Besides, we also evaluate FourLLIE on five unpaired datasets DICM \cite{lee2012contrast} (64 images), LIME \cite{guo2016lime} (10 images), MEF \cite{ma2015perceptual} (17 images), NPE \cite{wang2013naturalness} (85 images), and VV \footnote{https://sites.google.com/site/vonikakis/datasets} (24 images).

We implement FourLLIE in PyTorch and train it on an NVIDIA 3090 GPU for 0.5 days. The learning rate is initialized to $4.0\times10^{-4}$ and a multi-step scheduler is adopted. We choose Adam \cite{kingma2014adam} with momentum 0.9 as the optimizer. During the training, the input image is cropped to $384\times384$, and random rotation and flip augmentations are adopted. The batch size is set to 4 and the total training iterations are set to $2.0\times10^{5}$.

\begin{figure*}
    \centering
    \includegraphics[width=1.\textwidth]{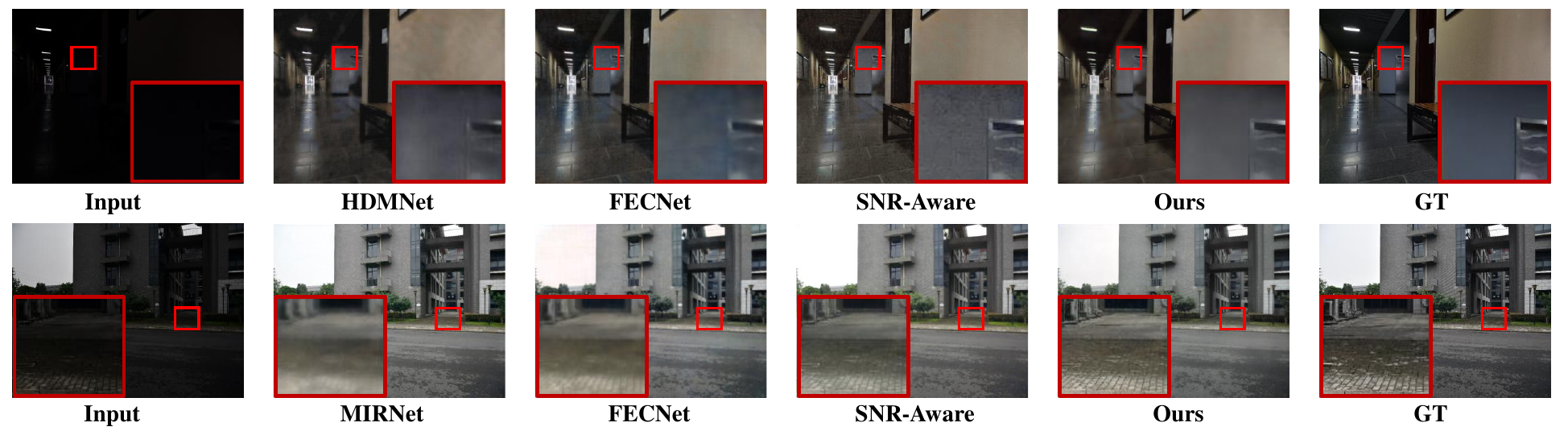}
     \vspace{-0.5cm}
    \caption{Qualitative comparison in LSRW-Huawei \cite{hai2023r2rnet} (the first row) and LSRW-Nikon \cite{hai2023r2rnet} (the second row). It can be seen that the proposed method reaches the best visual results. Note that we compare the proposed method with three benchmarks with the best PSNR values on the corresponding datasets. }
    \label{fig:comparison_LSRW}
    \vspace{-0.2cm}
\end{figure*}

\subsection{Comparison with State-Of-The-Arts \label{comparison_section}}

In this paper, we compare the proposed method with thirteen state-of-the-art LLIE methods, including traditional methods LIME \cite{guo2016lime}, MF \cite{fu2016fusion}, NPE \cite{wang2013naturalness}, SRIE \cite{fu2016weighted} and deep learning-based methods DRD\ cite{wei2018deep}, Kind \cite{zhang2019kindling}, Kind++ \cite{zhang2021beyond}, MIRNet \cite{zamir2020learning}, SGM \cite{yang2021sparse}, FECNet \cite{huang2022deep}, HDMNet \cite{liang2022learning}, SNR-Aware \cite{xu2022snr}, Bread \cite{guo2023low}. Note that all deep learning-based methods are trained on the same datasets with respective public codes.

\begin{table}[]
\caption{NIQE scores on DICM, LIME, MEF, NPE, and VV datasets. The best results are boldfaced and the second-best ones are underlined. Note that ``AVG'' represents the average values of the NIQE score on five datasets. All methods are pre-trained on LSRW-Huawei \cite{hai2023r2rnet}.}
 \vspace{-0.3cm}
    \label{tab:niqe}
    \resizebox{\columnwidth}{!}{
\renewcommand{\arraystretch}{1.} 
    \centering

    \begin{tabular}{c c c c c c c }
     \toprule
      Methods & LIME  &  VV & DICM &NPE &MEF &AVG \\
     \midrule
     KinD & 4.772 & 3.835 & 3.614 &  4.175   &4.819  &4.194 \\
     MIRNet  & 6.453 & 4.735 & 4.042 & 5.235  & 5.504 &5.101 \\
     SGM & 5.451 & 4.884 & 4.733 & 5.208  &  5.754 & 5.279\\
     FECNet  & 6.041 & 3.346  & 4.139 & 4.500  & 4.707 & 4.336 \\
     HDMNet & 6.403 & 4.462 &4.773 &5.108&  5.993 & 5.056 \\
     SNR-Aware& \underline{4.618} & \textbf{2.207} & \textbf{3.227} & \underline{3.975} & \underline{4.589} & \textbf{3.887}  \\
     Bread  & 4.717 & 3.304& 4.179 &4.160 &5.369 & 4.194 \\
     \midrule
     Ours & \textbf{4.402} &\underline{3.168}  & \underline{3.374} & \textbf{3.909} & \textbf{4.362} & \underline{3.907} \\
     \bottomrule
    \end{tabular}
    }
\vspace{-0.4cm}
\end{table}

\noindent\textbf{Quantitative comparison.}
We adopt Peak Signal-to-Noise Ratio (PSNR), Structural Similarity Index (SSIM) \cite{wang2004image}, and Learned Perceptual Image Patch Similarity (LPIPS) \cite{zhang2018unreasonable} as the evaluation matrices. LPIPS measures the distance between two images in high-level features. Generally, the higher PSNR, higher SSIM, and lower LPIPS represent that two images are more similar. 

As shown in Table \ref{tab:experiments_lol}, compared with recent methods, the proposed method reaches the best results in most of the cases, for the rest it almost achieves the second-best. Notably, compared with a recent Transformer based method SNR-Aware \cite{xu2022snr}, the proposed method achieves generally superior performance with only 0.03 $\%$ parameters due to the efficient global representation of the Fourier frequency information. Compared with a recent Fourier-based method FECNet \cite{huang2022deep}, the proposed method overall outperforms it with less parameters benefited from further use of the Fourier frequency information. Besides, our method also has competitive performance with UHDFour \cite{li2023embedding}, which has 21.78dB PSNR and 0.87 SSIM on LOL-Real \cite{yang2021sparse} but with 17.54M parameters.

Besides, we measure the naturalness image quality evaluator (NIQE) score on five unpaired datasets. The images with the lower NIQE scores represent the higher naturalness image quality. Table \ref{tab:niqe} presents the results of the NIQE evaluation. It can be seen that the proposed method outperforms most existing LLIE methods. Although SNR-Aware \cite{xu2022snr} reaches competitive performance in the NIQE evaluation, it has much more parameters than the proposed method (see Table \ref{tab:experiments_lol}).

\begin{table}
\caption{Quantitative comparison on the SICE \cite{cai2018learning}. The best results are boldfaced and the second-best ones are underlined. Note that the results of other methods are collected from \cite{huang2022deep}.}
    \label{tab:SICE}
    \vspace{-0.3cm}
\renewcommand{\arraystretch}{1.} 
    \centering
    \resizebox{\columnwidth}{!}{
      \begin{tabular}{ c c c c c c c   }
         \toprule
    \multirow{2}*[-0.2cm]{Methods} &  \multicolumn{2}{c}{Under}&  \multicolumn{2}{c}{Over} &   \multicolumn{2}{c}{Average}     \\
     \cmidrule(r){2-7}
     & PNSR & SSIM & PNSR & SSIM & PNSR & SSIM  \\
        \midrule
     DRD \cite{wei2018deep} & 12.94&0.5157 &   12.87 &0.5252  &12.90 & 0.5212 \\
        DPED \cite{ignatov2017dslr} & 16.83 & 0.6133 &  7.99 & 0.4300& 12.41 & 0.5217 \\
        DRBN \cite{yang2020fidelity}&17.96 & \underline{0.6767} &  17.33 &0.6828 & 17.65 & 0.6798\\
        SID \cite{chen2018learning} & 19.51&0.6635  &  16.76 &0.6444 & 18.15 & 0.6540  \\
        MSEC \cite{Afifi_2021_CVPR}&19.62 & 0.6512 &17.59   & 0.6560& 18.58 & 0.6536\\
        CMEC \cite{nsamp2018learning}&17.68 & 0.6592 & 18.17  & 0.6811&  17.93& 0.6702 \\
        FECNet \cite{huang2022deep} & \textbf{22.01}& 0.6737 & \underline{19.91}  & \underline{0.6960}&\textbf{20.96} & \underline{0.6849 }\\
      
     \midrule
     \multicolumn{1}{c}{Ours} & \underline{21.36} & \textbf{0.6793}  &\textbf{ 20.14}  & \textbf{0.7143}  & \underline{20.75}  & \textbf{0.6968}\\
     \bottomrule
     
    \end{tabular}}
    \vspace{-0.4cm}
\end{table}

\begin{table*}
\caption{Ablation study.
The best results are boldfaced and the second-best ones are underlined.}
\vspace{-0.3cm}
    \label{tab:ablation}
       \resizebox{\textwidth}{!}{
\renewcommand{\arraystretch}{1.} 
    \centering
      \begin{tabular}{c c c c c c c c c c c c c c}
         \toprule

     \multirow{2}{*}{Methods} &\multicolumn{3}{c}{LOL-Real \cite{yang2021sparse}}&\multicolumn{3}{c}{LOL-Synthetic \cite{yang2021sparse}}&\multicolumn{3}{c}{LSRW-Huawei \cite{hai2023r2rnet}}&\multicolumn{3}{c}{LSRW-Nikon \cite{hai2023r2rnet}}&\multirow{2}{*}{\makecell[c]{\#Param\\(M)}} \\
      \cmidrule(r){2-4} \cmidrule(r){5-7} \cmidrule(r){8-10} \cmidrule(r){11-13}
     &    PSNR   &   SSIM  &   LPIPS  &   PSNR    &   SSIM &   LPIPS &   PSNR     &   SSIM   &   LPIPS &  PSNR   &   SSIM &   LPIPS   \\
        \midrule
      Ours w/o F & 21.01 & \underline{0.8398}  & \underline{0.0571} & \underline{24.14} & \textbf{0.9196} & \textbf{0.0297} & 20.61 & 0.5986 & 0.1928 & 17.24 & 0.4866 & 0.1898 & \underline{0.09}\\
      Ours w/o S & 20.20 & 0.7128  &0.1199 & 17.70 & 0.7156 & 0.1423 & 17.25 & 0.5358 & 0.1976 & 15.17 & 0.4328 & 0.1714 & \textbf{0.03} \\
      Ours w/o SNR & 20.52 & 0.8291  &0.0638 & 22.39 & 0.9009 & 0.1470 & 20.61 & 0.6020 & 0.2012 & 17.19 & 0.4889 & \textbf{0.1269} & 0.12\\
      Ours w/o $\mathcal{L}_{s1}$ & \underline{21.29} & 0.8282  &0.0634 & 22.94 & 0.9036 & 0.0443 & \underline{20.96} & 0.6050 & 0.2303 & 16.93 & 0.4761 &\underline{ 0.1464} & 0.12 \\
      Ours w/o $\mathcal{L}_{vgg}$  & 19.61 & 0.7874  & 0.1050 & 21.54 & 0.8795 & 0.0501 & 20.70 & \underline{0.6151} & \underline{0.1745} & \underline{17.27} & \underline{0.5020} & 0.2152 & 0.12 \\
      \midrule
      Ours & \textbf{22.34} & \textbf{0.8468}  & \textbf{0.0511} & \textbf{24.65} & \underline{0.9192} & \underline{0.0389} & \textbf{21.30} & \textbf{0.6220} & \textbf{0.1719} & \textbf{17.82} & \textbf{0.5036} & 0.2150  & 0.12\\
     \bottomrule
     
    \end{tabular}}
\vspace{-0.4cm}
\end{table*}

\noindent\textbf{Qualitative comparison.}
We first present the qualitative comparison of LOL-Real and LOL-Synthetic in Fig. \ref{fig:comparison_LOL}, where the first row is the results of LOL-Real and the second row is the results of LOL-Synthetic. We select three methods with the best PSNR values. It can be seen that the proposed method reaches the best visual results both in detail and lightness. Even for some complex textures, the proposed method can recover well (see the first row of Fig. \ref{fig:comparison_LOL}). 

Fig. \ref{fig:comparison_LSRW} presents the comparison results on LSRW-Huawei (first row) and LSRW-Nikon (second row). It can be seen that the results of the proposed method have fewer noise and clearer details than other methods. 

\begin{figure*}
    \centering
    \includegraphics[width=1.\textwidth]{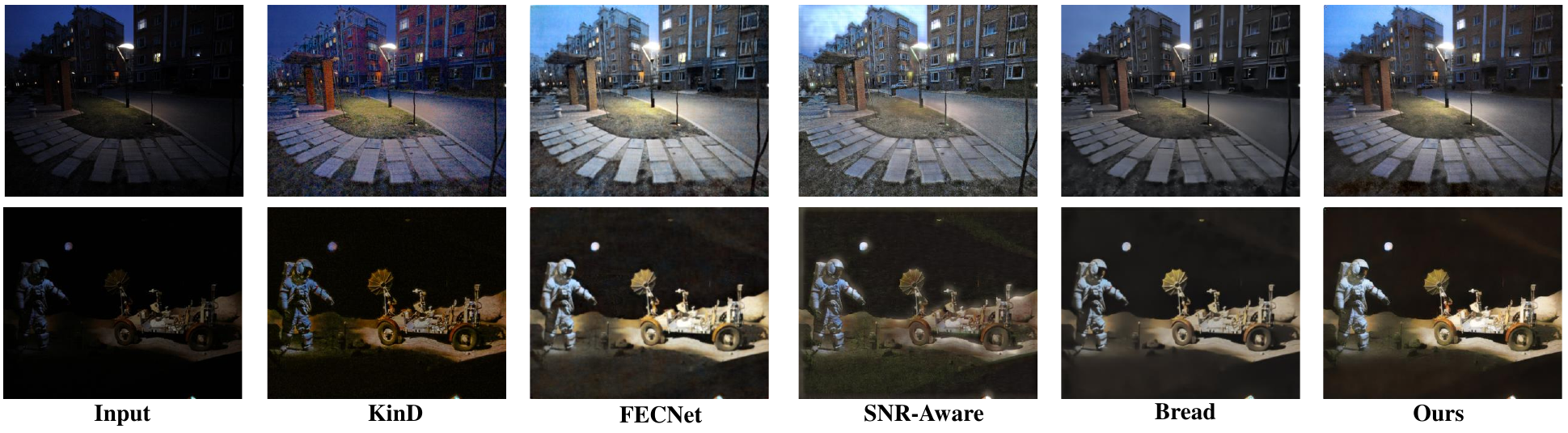}
     \vspace{-0.7cm}
    \caption{Qualitative comparison in unpaired datasets (the first row is DICM \cite{lee2012contrast} and the second row is LIME \cite{guo2016lime}). It can be seen that the proposed method reaches the best visual results. Note that we compare the proposed method with four benchmarks with the best average NIQE values in Table \ref{tab:niqe}. }
    \label{fig:unpair_visual}
    \vspace{-0.4cm}
\end{figure*}

Besides, we also present the visual comparison on unpaired datasets. Fig. \ref{fig:unpair_visual} presents the results of DICM (first row) and LIME (second row). We can see that the proposed method has best visual results, which demonstrates the decent generalization of the proposed method. The rest visual results can be found in supplementary material.

\subsection{Extension on Exposure Correction. \label{extension}}
In this work, we find the lightness of low-light images can be improved by enlarging the magnitude of the amplitude component in the Fourier space. Alternatively, the lightness also can be depressed by reducing the magnitude of amplitude. Therefore, we extend FourLLIE to the exposure correction task and evaluate it in the SICE \cite{cai2018learning} dataset, which contains 512 over-/normal- exposure image pairs and 512 under-/normal- exposure pairs for training, and 60 image pairs for testing. 
Note that the frequency stage should estimate an additional amplitude transform map for lightness depression (details can be seen in supplementary materials). The results in Table \ref{tab:SICE} show that the proposed method outperforms existing methods in PSNR and SSIM. It demonstrates the potential of the proposed method in exposure correction.

\subsection{Ablation Study} 

FourLLIE utilizes the Fourier frequency information to enhance the low-light images through two stages: frequency stage and spatial stage. In this subsection, we conduct the ablation study with five different settings to verify the effectiveness of the proposed designs and adopted loss functions.
\textbf{1)} ``Ours w/o F'' removes the frequency stage. \textbf{2)} ``Ours w/o S'' removes the spatial stage. \textbf{3)} ``Ours w/o SNR'' removes SNR-based frequency and spatial interaction and replaces it with the feature concatenation. \textbf{4)} ``Ours w/o $\mathcal{L}_{s1}$'' removes the $\mathcal{L}_{s1}$. \textbf{5)} ``Ours w/o $\mathcal{L}_{vgg}$'' removes the perceptual loss $\mathcal{L}_{vgg}$ of $\mathcal{L}_{s2}$. Table \ref{tab:ablation} presents the results on both four datasets. Compared with all ablation settings, our full setting almost reaches the best PNSR, SSIM, and LPIPS. It demonstrates the effectiveness of the proposed designs and adopted loss functions.

\section{Conclusion}
Inspired by previous Fourier-based LLIE methods, in this work, we further explore the properties of the Fourier frequency information and propose a new Fourier-based LLIE method. We first analyze the relationship between the amplitude component and lightness and conclude that lightness can be improved by enlarging the magnitude of the amplitude component. Then, we find the Fourier frequency information has nice global properties and is efficient to extract, since it does not introduce massive parameters of neural networks. Based on the above observations, we design a two-stage architecture FourLLIE, which first estimates the amplitude transform map to improve the lightness in the frequency stage. Then, the SNR map is introduced to accomplish frequency and spatial interaction in the spatial stage and to recover the detail. Benefiting from the effectiveness of Fourier frequency information, FourLLIE outperforms existing SOTA LLIE methods with a lightweight architecture. 

Recently, many related works have demonstrated the enormous potential of the Fourier frequency information in low-/high- level computer vision tasks. In the future, we will explore more properties of the Fourier frequency information to make the proposed method adaptive for more diverse degradation.

\begin{acks}
This work was supported by the National Natural Science Foundation of China under Grant No. 62071500. Supported by Sino-Germen Mobility Programme M-0421. 
\end{acks}

\balance
\bibliographystyle{ACM-Reference-Format}
\bibliography{sample-sigconf}

\clearpage
\appendix
\section{Detailed experiment settings in section 3.2}
\begin{figure*}
  \centering

  \captionsetup[subfloat]{labelformat=empty}
  
  \subfloat[ 
  ]{
    \includegraphics[width=1.\textwidth]{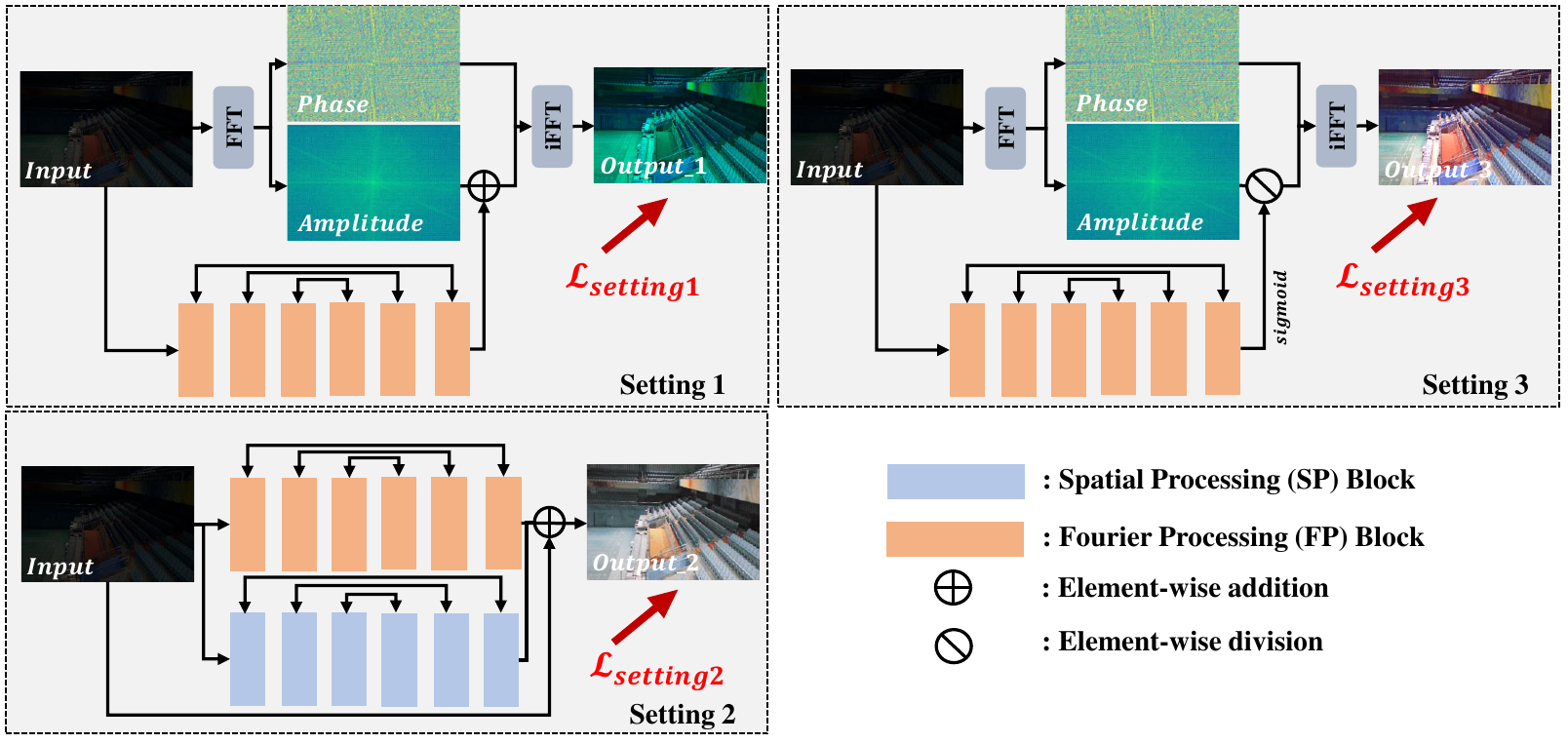}
  }
  \vspace{-0.3cm}
  \caption{Detailed experiment settings in Section 3.2. Note that considering the input and output of setting 2 are in spatial space, we add a spatial branch for better convergence. }
    \label{fig:diff_settings}
    \vspace{-0.3cm}
\end{figure*}

In Section 3.2, we conduct an experiment to demonstrate the effectiveness of estimating the amplitude transform map. In this part, we present the detailed implementations of this experiment. As shown in Fig. \ref{fig:diff_settings}, setting 1 predicts the amplitude component directly (we add a residual connection for faster convergence), this process can be expressed by:

\begin{equation}
\begin{split}
&A_{res} = NN(input) \\
A_{in} = \mathcal{A}&(\mathcal{F}(input)), \quad P_{in} = \mathcal{P}(\mathcal{F}(input))\\
&A_{out} = A_{in} + A_{res}  \\
R_{out} = A_{out} &\times cos(P_{in}) \quad I_{out} = A_{out} \times sin(P_{in})\\
&Output\_1 = \mathcal{F}^{-1}(R_{out} + I_{out}j)
\end{split} 
\end{equation}
where $NN$ represents the neural networks in Fig. \ref{fig:diff_settings}. Then, the constraint of setting 1 $\mathcal{L}_{setting1}$ is expressed as:
\begin{equation}
    \mathcal{L}_{setting1} = \left | \left | \mathcal{A}(\mathcal{F}(Output\_2))-\mathcal{A}(\mathcal{F}(GT)) \right |  \right | _{2}
\end{equation}
where GT is the ground truth. 

Setting 2 predicts the enhanced image directly (residual connection is also adopted) and employs constrain on the amplitude component of the predicted enhanced image. This process can be expressed by:
\begin{equation}
    Output\_2 = NN(input)
\end{equation}
then, the constraint of setting 1 $\mathcal{L}_{setting1}$ is expressed as:
\begin{equation}
    \mathcal{L}_{setting2} = \left | \left | \mathcal{A}(\mathcal{F}(Output\_2))-\mathcal{A}(\mathcal{F}(GT)) \right |  \right | _{2}
\end{equation}

Setting 3 estimates the amplitude transform map based on the observation that the magnitudes of the amplitude component can reflect the magnitudes of the lightness. This process can be expressed by:
\begin{equation}
\begin{split}
&M = sigmoid(NN(input)) \\
A_{in} = \mathcal{A}&(\mathcal{F}(input)), \quad P_{in} = \mathcal{P}(\mathcal{F}(input))\\
&A_{out} = A_{in} / (M + \epsilon ) \\
R_{out} = A_{out} &\times cos(P_{in}) \quad I_{out} = A_{out} \times sin(P_{in})\\
&Output\_3 = \mathcal{F}^{-1}(R_{out} + I_{out}j)
\end{split} 
\end{equation}
where $M$ is the amplitude transform map, $\epsilon$ is set to $1e-8$ for avoiding zero-division. Then, the constraint of setting 3 $\mathcal{L}_{setting3}$ is expressed as:
\begin{equation}
    \mathcal{L}_{setting3} = \left | \left | \mathcal{A}(\mathcal{F}(Output\_3))-\mathcal{A}(\mathcal{F}(GT)) \right |  \right | _{2}
\end{equation}

\begin{figure}
    \centering
    \includegraphics[width=1.\columnwidth]{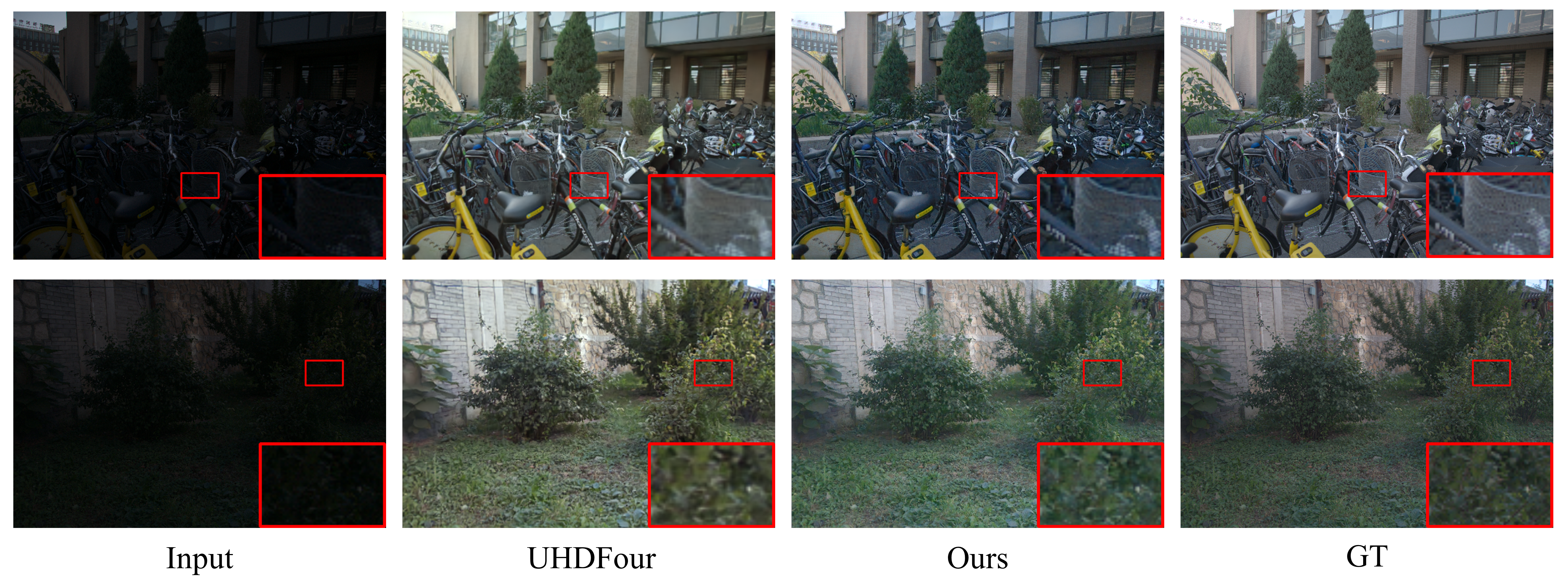}
     \vspace{-0.3cm}
    \caption{Visual comparison with UHDFour \cite{li2023embedding}. }
    \label{fig:uhd_vis_cmp}
    \vspace{-0.3cm}
\end{figure}

\begin{figure*}
  \centering

  \captionsetup[subfloat]{labelformat=empty}
  
  \subfloat[
  ]{
    \includegraphics[width=1.\textwidth]{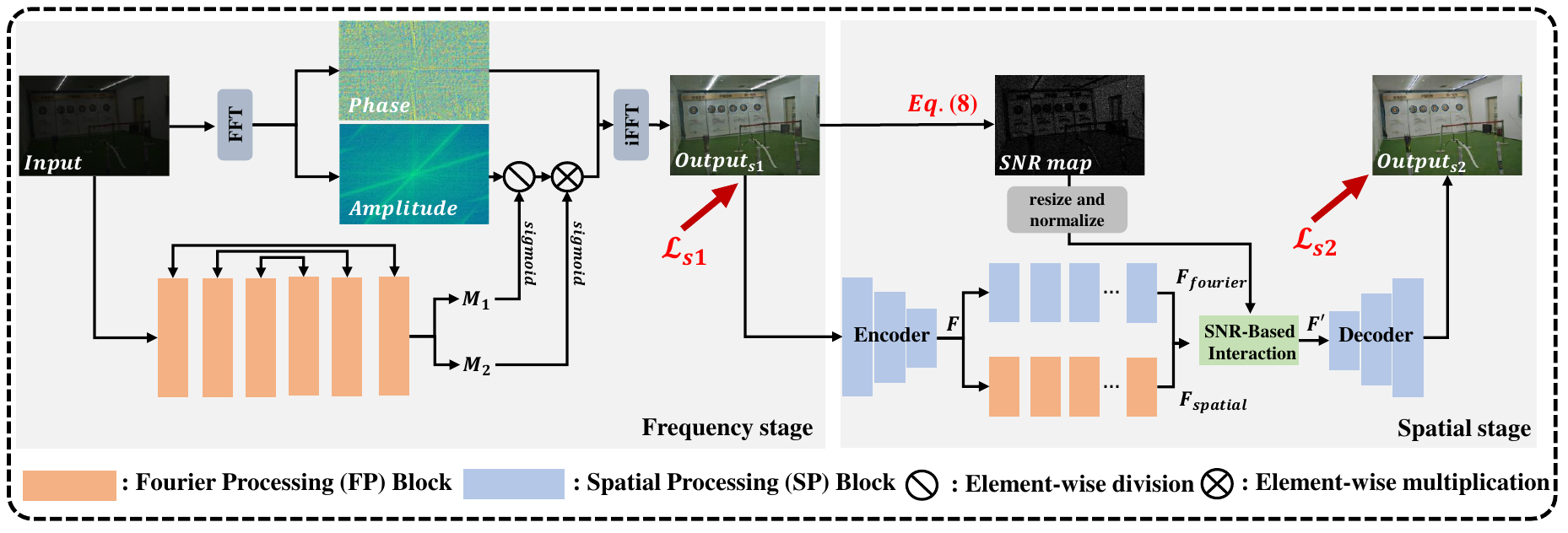}
  }
  \vspace{-0.3cm}
  \caption{The variant of FourLLIE for exposure correction. It estimates an additional amplitude transform map to depress the magnitudes of amplitude to accomplish the lightness depression. }
    \label{fig:pipline_EC}
    \vspace{-0.3cm}
\end{figure*}

\section{variant for exposure correction}
To accomplish over-exposure depression, FourLLIE needs to be able to reduce the magnitudes of amplitude components. As shown in Fig. \ref{fig:pipline_EC}, we predict two amplitude transform maps, where one is responsible for improving the lightness and another is responsible for depressing the lightness, to accomplish exposure correction.

\section{comparison with UHDFour}

We compare the proposed method with a recent Fourier-based LLIE method UHDFour \cite{li2023embedding}. As shown in Table \ref{tab:quan_comparison} and Fig. \ref{fig:uhd_vis_cmp}, the proposed method reaches overall better performance.

\begin{table}
\caption{Quantitative comparison with UHDFour \cite{li2023embedding}. Note that UHDFour (8×) represents UHDFour with 8× down-sampling in the low-resolution network (LRNet) and is suitable for high-resolution images, while UHDFour (2×) represents UHDFour with 2× down-sampling in LRNet and is suitable for low-resolution images.}
\vspace{-0.2cm}
    \label{tab:quan_comparison}
       \resizebox{\columnwidth}{!}{
\renewcommand{\arraystretch}{1.1} 
    \centering
      \begin{tabular}{c c c c c c c c c c}
         \toprule

     \multirow{2}{*}{Methods} &\multicolumn{2}{c}{LOL-Real \cite{yang2021sparse}}&\multicolumn{2}{c}{LOL-Synthetic \cite{yang2021sparse}}&\multicolumn{2}{c}{LSRW-Huawei \cite{hai2023r2rnet}}&\multicolumn{2}{c}{LSRW-Nikon \cite{hai2023r2rnet}}&\multirow{2}{*}{\makecell[c]{\#Param\\(M)}} \\
      \cmidrule(r){2-3} \cmidrule(r){4-5} \cmidrule(r){6-7} \cmidrule(r){8-9}
     &    PSNR   &   SSIM    &   PSNR    &   SSIM &   PSNR     &   SSIM   &     PSNR   &   SSIM   \\
        \midrule
      UHDFour(8×) & 20.91 &0.76  & 22.70 & 0.87& 20.64 & 0.58  & 17.72 & 0.48 & 17.54\\
      UHDFour(2×) & \underline{21.78} & \textbf{0.87} & \underline{23.74} & \underline{0.90}  & \underline{20.88} & \underline{0.61} & \underline{17.57} & \textbf{0.51} &  17.54\\
      \midrule
      Ours & \textbf{22.34} & \underline{0.85}  & \textbf{24.65} & \textbf{0.92} & \textbf{21.30} & \textbf{0.62}  & \textbf{17.82} & \underline{0.50} & 0.12\\
     \bottomrule
     
    \end{tabular}}
\vspace{-0.2cm}
\end{table}

\end{document}